\documentclass{article}
\usepackage{spconf,amsmath,graphicx}

\usepackage[ruled]{algorithm2e}
\usepackage{booktabs}
\usepackage{multirow}
\usepackage{tabularx}
\usepackage{graphicx}
\usepackage{threeparttable}
\usepackage{makecell}
\usepackage{amsmath,amsthm,amssymb}
\usepackage{amsfonts}
\usepackage{mathrsfs}
\usepackage{blindtext}
\usepackage{subfigure}

\usepackage{lipsum}

\title{Advancing Underwater Acoustic Target Recognition via Adaptive Data Pruning and Smoothness-inducing Regularization}
\name{Yuan Xie\textsuperscript{\rm 1} \textsuperscript{\rm 2}, Tianyu Chen \textsuperscript{\rm 3} \textsuperscript{\rm 4},Ji Xu\textsuperscript{\rm 1} \textsuperscript{\rm 2}}

\address{
\textsuperscript{\rm 1}Key laboratory of Speech Acoustics and Content Understanding\\ Institute of Acoustics, Chinese Academy of Sciences \\
\textsuperscript{\rm 2}University of Chinese Academy of Sciences \\
\textsuperscript{\rm 3}BDBC, Beihang University, China
\textsuperscript{\rm 4}SKLSDE, Beihang University
}

\begin{document}
%
\maketitle
\begin{abstract}
Underwater acoustic recognition for ship-radiated signals has high practical application value due to the ability to recognize non-line-of-sight targets. However, due to the difficulty of data acquisition, the collected signals are scarce in quantity and mainly composed of mechanical periodic noise. According to the experiments, we observe that the repeatability of periodic signals leads to a double-descent phenomenon, which indicates a significant local bias toward repeated samples. To address this issue, we propose a strategy based on cross-entropy to prune excessively similar segments in training data. Furthermore, to compensate for the reduction of training data, we generate noisy samples and apply smoothness-inducing regularization based on KL divergence to mitigate overfitting. Experiments show that our proposed data pruning and regularization strategy can bring stable benefits and our framework significantly outperforms the state-of-the-art in low-resource scenarios.


\end{abstract}
\section{Introduction}
\label{sec:intro}
Acoustic target recognition aims to automatically recognize non-line-of-sight targets\cite{robertson1995artificial}. Long detection range, reassuring concealment, and low deployment cost make it irreplaceable in practical application\cite{heupel2006automated}. With the advent of the deep learning era\cite{lecun2015deep} era, data-driven neural networks have achieved overwhelming success and have become the predominant method for acoustic recognition\cite{gombos2019acoustic}.

\renewcommand{\floatpagefraction}{.8}
\begin{figure}
    \centering
    \subfigure[To mechanically repeated signals, the spectrograms of the 30-second segments are highly similar when the working condition is stable.]{
        \begin{minipage}[b]{0.42\textwidth}
        \includegraphics[width=\linewidth]{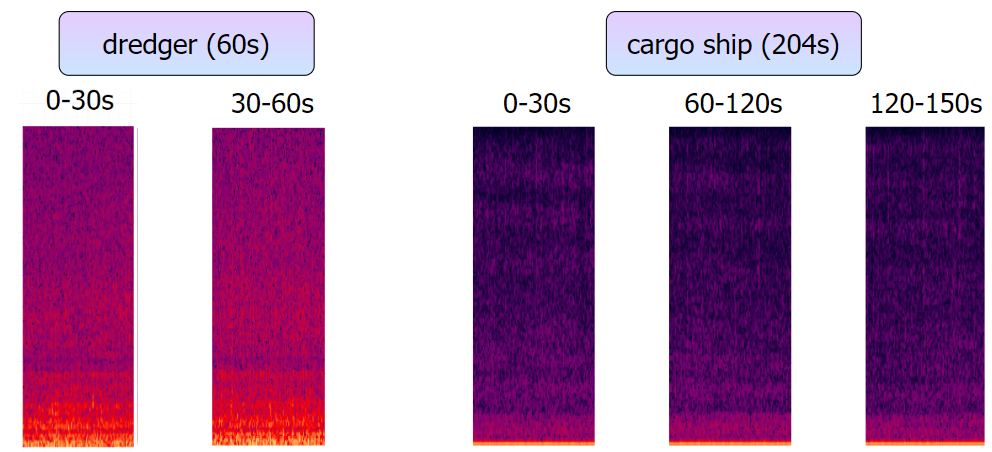}
        \end{minipage}}

    \subfigure[Test loss curve of the model trained on ship-radiated signals. In addition to overfitting, the baseline model suffers from a double-descent problem.]{
        \begin{minipage}[b]{0.48\textwidth}
        \includegraphics[width=\linewidth]{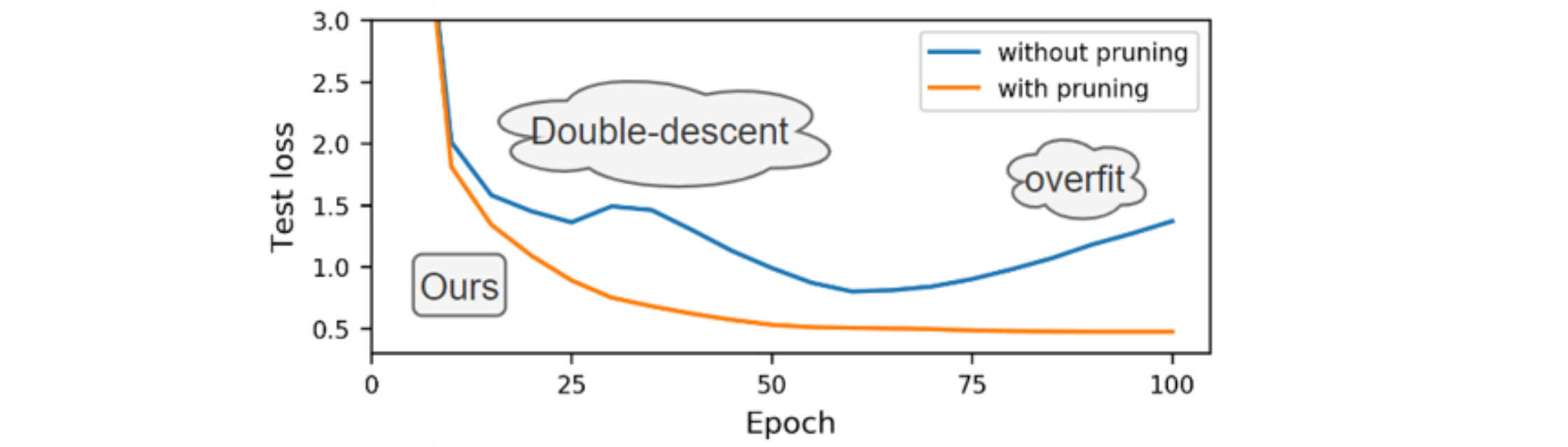}
        \end{minipage}}
    \caption{Two core issues of acoustic target recognition in natural environments and the optimization brought by our proposed methods.}
    \label{fig: issue}
    \vspace{-5px}
\end{figure}

Due to the challenge of data acquisition, the ship-radiated signals are usually scarce in quantity\cite{xie2022underwater} and mainly composed of mechanical periodic repetitive noise\cite{perrone1975analysis}, such as propeller cavitation noise, or rhythm modulation noise caused by diesel piston movement. Due to the varying duration of the collected signals (take the databases used in this paper as an example, the duration of the signals ranges from 9 seconds to 31.5minutes), we need to split long signals into short-time segments for training to preserve sufficient time and frequency resolution and limit the feature dimensions. However, when the ships' working condition is stable, the short-time segments may have high degrees of similarity due to periodic repetitive noise. Fig.\ref{fig: issue}(a) shows the time-frequency spectrograms of different 30-second segments from two selected signals, which exhibit high similarity. It inevitably introduces similar training samples to the training data. Previous research\cite{hernandez2022scaling} has shown that the double-descent phenomenon occurs when the model is biased towards repeated data. It causes models to fall into local bias and suffer from performance degradation\cite{hernandez2022scaling, stowell2015acoustic}. In Fig.\ref{fig: issue}(b), we verify that our similar signal segments also bring a significant double descent phenomenon during the training stage. To the best of our knowledge, this problem has not yet been explored in the field of acoustic recognition of ship-radiated signals.

In addition to the double-descent problem, we observe that the recognition model also severely suffers from overfitting (see Fig.~\ref{fig: issue}(b)). While mitigating overfitting has been a hot topic for many years, data augmentation is currently the most prevalent method. It alleviates overfitting by generating virtual samples to simulate real data distributions. Some researchers analyze the acoustic characteristics of signals and design manual perturbations to generate noisy samples\cite{chen2019integrating, cui2015data}, while some researchers diversify data distributions by generating virtual samples through generative adversarial networks (GANs)\cite{gao2020recognition, yang2020gan}. However, in complex acoustic scenarios such as marine environment\cite{xie2022adaptive, ren2022ualf}, manual data augmentation and GANs may not consistently lead to improvements due to inevitable deviations\cite{gong2021eliminate} between simulations and real marine environments. If data augmentation pushes supplementary data further from the real distribution, the accuracy of recognition models may even degrade. Our experimental results also confirm this issue (see Section 4.1). Therefore, novel techniques that are less affected by the quality of noisy samples are needed.

To solve the issues discussed above, our work presents an effective and efficient training pipeline consisting of two key components. First, we propose an adaptive data pruning strategy to prune mitigate similar data and eliminate the  double-descent phenomenon. We add a linear pruning layer dedicated to calculating the pruning scores to realize dynamic training data pruning. Furthermore, to alleviate the overfitting caused by the reduction of training data, we propose smoothness-inducing regularization inspired by local shift sensitivity from robust statistics\cite{huber2011robust}. To ensure stable performance improvement, we use noisy samples generated from Gaussian white noise to calculate the regularization term, rather than as supplementary training data. The regularization constraints result in a smoother decision boundary for the model, and it can minimize the risk of bias toward a non-representative data distribution  by excluding noisy samples from directly calculating the loss.

We evaluate our proposed methods on both low-resource and medium-resource databases, verifying that they consistently outperform state-of-the-art. Notably, on low-resource data, the accuracy rate improves significantly from 85.48\% (SOTA) to 88.79\%. Besides, we demonstrate that our data pruning strategy could improve the training efficiency by up to 65\% while steadily benefiting the model performance.

\vspace{-2mm}
\section{Method}

\vspace{-2mm}

\subsection{Adaptive Data Pruning}

\begin{figure}[t]
    \centering
    \includegraphics[width=\linewidth]{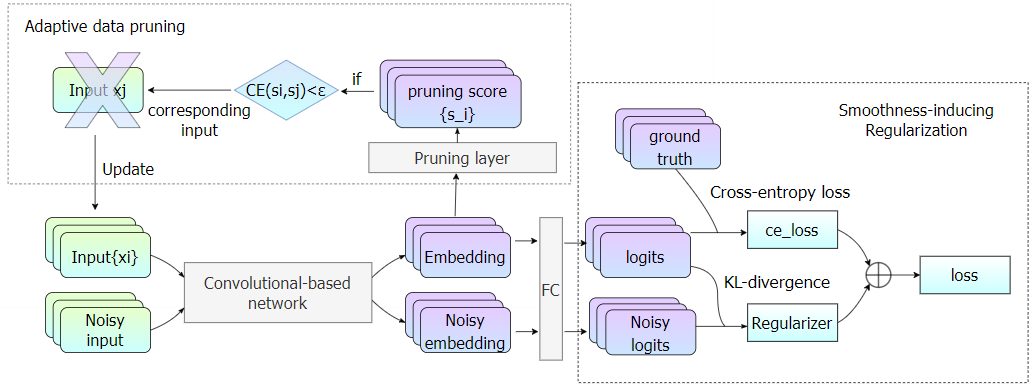}
    \caption{Pipeline for our proposed data pruning and regularization. ``CE'' represents cross-entropy, ``FC'' represents fully-connected layer.}
    \label{fig:pipeline}
\vspace{-1mm}
\end{figure}

According to our analysis in Section 1, it is necessary to mitigate data similarity and eliminate the double-descent phenomenon. In this work, we propose an adaptive cross-entropy-based data pruning strategy to realize dynamic data pruning of training data. Denote the signals as $x$ and corresponding label as $y$, the $n$ training samples in a batch could be represented as \{($x_i,y_i$)\} ($i = 1, 2 ...n$). As shown in Fig.~\ref{fig:pipeline}, the front-end convolutional-based network is used to extract the spectrogram embeddings. During the first $\tau$ epochs (by default, $\tau=10$), we do not perform data pruning as the model is not adequately trained at this stage. We then add a linear pruning layer to calculate the pruning score $s_i$ for each input. The purpose of this pruning layer is to reduce the dimensionality of embeddings and facilitate subsequent cross-entropy operations. Subsequently, we calculate the cross-entropy of all $s_i$ in a batch with each other. If the cross-entropy between $s_i$ and $s_j$ is below the threshold $\varepsilon$, we randomly prune corresponding $x_i$ or $x_j$ and prohibit it from participating in subsequent epochs. It is worth noting that the threshold should be set to a very small value (e.g., 1e-5) to avoid aggressive pruning. Furthermore, to prevent the model from forgetting previously learned knowledge, we employ an early stopping strategy, which terminates the training process when the validation loss does not decrease over several epochs (by default, 10 epochs).

\subsection{Smoothness-inducing Regularization}
The first step of smoothness-inducing regularization is to generate noisy samples. We  could apply relevant perturbations (adding background noise, adding pulse, or spectral shifting) based on possible acoustic interferences to generate noisy samples. In this work, we simply add 5$\sim$30dB Gaussian white noise as the perturbation. As illustrated in Fig.~\ref{fig:pipeline}, for each epoch, we randomly add perturbations to input signals $x_i$ to dynamically generate noisy samples ($\tilde{x_i},y_i$). During training, the model $f(\cdot)$ takes both ($x_i,y_i$) and ($\tilde{x_i},y_i$) as input, and outputs the logits $z_i=f(x_i)$ and $\tilde{z_i}=f(\tilde{x_i})$ after the forward propagation. Inspired by previous work\cite{jiang2019smart}, we apply Kullback-Leibler (KL) divergence to measure the difference between the logits of the raw and noisy samples and employ it as the basis for our regularization term. Our optimization goal is to minimize the loss function:

\begin{equation}
\begin{aligned}
    \mathcal L
    &=\mathcal L_{task} + \alpha \mathcal R \\
    &=\mathcal L_{CE}(z_i,y_i) + \alpha
    (\mathcal L_{KL}(z_i,\tilde{z_i})+ \mathcal L_{KL}(\tilde{z_i},z_i)) 
\end{aligned}
\end{equation}

where $\mathcal L_{CE}$ refers to the cross-entropy loss and $\alpha$ is an adjustable parameter to control the regularization weight. $\mathcal L_{KL}$ represents the regularization term based on KL divergence. We feed the logits $z_i$ and $\tilde{z_i}$ into the softmax function and get the normalized probability distribution $p_i$ and $\tilde{p_i}$. The KL divergence-based term could be formulated as:


\begin{equation}
\begin{aligned}
    \mathcal L_{KL}(z_i,\tilde{z_i}) 
    &= \mathcal D_{KL}(p_i\Vert \tilde{p_i})
    &=  \frac{1}{n}\sum^{n}_{i=1} p_ilog\frac{p_i}{\tilde{p_i}}
\end{aligned}
\vspace{-2mm}
\end{equation}

The regularization term in Equation 2 gets minimized when the model makes similar predictions for $x_i$ and its noisy neighbors $\tilde{x_i}$. Since noisy samples are not involved in the computation of the loss with ground truth, the model will minimize the risk of aggravating misjudgment even if noisy samples deviate from real-world scenarios. This property makes the decision boundary of the model smoother and endows the model with greater stability and resilience to perturbations.

\section{Experiment Setup}

\subsection{Benchmark dataset}
In this work, we select two underwater ship-radiated noise datasets at different scales to verify the effectiveness of our methods on different resource levels.

\textbf{Shipsear}\cite{santos2016shipsear} (low-resource database): 
It consists of 3 hours, 90 recordings of ship and boat sounds. We conduct the 9-class recognition task (dredger, fish boat, motorboat, mussel boat, natural noise, ocean liner, passenger ship, roro ship, sailboat) on Shipsear.

\textbf{Deepship}\cite{irfan2021deepship} (medium-resource database):
It consists of 47-hour underwater recordings of 265 different ships belonging to four classes (cargo, passenger ship, tanker, and tug).

\subsection{Baseline Methods}

\textbf{Acoustic Features}: Prior studies have proven that the time-frequency-based spectrogram can achieve superior recognition performance\cite{ren2022ualf}. We perform the short-time Fourier transform to get STFT spectrograms and the constant Q transform to get CQT spectrograms. Additionally, Mel filter banks are employed to derive Mel spectrograms based on STFT spectrograms.

\textbf{Regularization Methods}: We implement the manual data augmentation and GANs\cite{yang2020gan} as strong baselines. For fair comparisons, we use the same perturbations (adding 5$\sim$30dB Gaussian white noise) for data augmentation and our smoothness-inducing regularization.

\subsection{Parameters setting}
For acoustic feature extraction, we uniformly take 50ms as the frame length and 25ms as the frame shift. Audio files are split into segments with 30s length and 15s overlapping. The number of filter banks is set to 300. During training, We use the Adam optimizer\cite{kingma2014adam} and cosine scheduler with warmup (5 epochs). The learning rate is set to 5e-4 for all experiments. The max training epoch is set to 100.

\section{Results And Discussion}
For our reported results, we take accuracy as the metric and count results on the 30-second segment level. Besides, we take ResNet\cite{he2016deep} with multi-head attention as the backbone according to our preliminary experiment. It changes the global average pooling layer into an attention pooling mechanism, where a ``QKV'' attention is implemented\cite{Wang2018NonlocalNN}.

\subsection{Main results}
\begin{table}[t]
  \centering
  \caption{Performance on underwater acoustic benchmarks. ``aug'' represents augmentation and ``smooth-reg'' represents smoothness-inducing regularization. The underlined results show that the regularization method leads to performance degradation.}
  \resizebox{\linewidth}{!}{
  \begin{threeparttable}
    \begin{tabular}{llcccc}
    \toprule
    & & \multicolumn{2}{c}{Shipsear} & \multicolumn{2}{c}{DeepShip}\\
    \cmidrule(lr){3-4}\cmidrule(lr){5-6}
    Features &Methods & prune($\times$)& prune($\checkmark$)  & prune($\times$)& prune($\checkmark$)\\
    \midrule
    STFT spec& -  & 75.24 & 79.59 & 74.68&75.88 \\
    &GANs\cite{yang2020gan} & 76.05 & 82.08 & 75.98& 76.62\\
    &Manual data aug & 78.45 & 81.41 & 75.07& 76.62\\
    &Smooth-reg & 81.90  & 86.78 & 76.38& 78.11\\
    \hline

    CQT spec& -  &73.33 &  75.91 & 77.82&  79.51 \\
    &GANs & 73.43 &  76.52  & \underline{77.12}& \underline{79.48}  \\
    &Data aug & 73.41 &  76.24  & 77.86& \underline{78.99} \\
    &Smooth-reg & 75.86  &  78.87  & 78.25& \textbf{80.19} \\
    \hline

    Mel spec& -  & 77.14 &  80.41 & 74.85& 76.01 \\
    &GANs & 77.68 &  82.11 & 75.79& 76.29 \\
    &Data aug & 77.72 & 82.40  & 75.27& \underline{75.85} \\
    &Smooth-reg & 82.76  & \textbf{88.79}  & 77.05& 78.77 \\
    \hline
    SOTA for Shipsear  &Fbsp Wavelet\cite{xie2022adaptive}& 85.48  & -  & 77.09 & - \\
    Benchmark for DeepShip {1}
    &SCAE\cite{irfan2021deepship}& -  & -  & 77.53 & - \\
    
    \bottomrule
    \end{tabular}
    \begin{tablenotes}
        \footnotesize
        \item[1] Since the DeepShip does not have an official train/test split, a fair SOTA is unavailable. We use the best result in~\cite{irfan2021deepship} as a benchmark.
    \end{tablenotes}
    \end{threeparttable}}
  \label{tab:main}%
  \vspace{-3mm}
\end{table}%

Table~\ref{tab:main} presents the results of our  experiments on data pruning and regularization strategies based on three acoustic features on the Shipsear and Deepship datasets. First, we observe that adaptive data pruning can consistently improve performance for various acoustic features and datasets. These results further demonstrate the importance of addressing the double-descent issue. We also find that data pruning works particularly well on low-resource Shipsear, which can bring up to a 6.03\% performance improvement. This is because the scarcity of data causes the model to be more likely to memorize repeated data and fall into local bias.

Regarding regularization methods, our baselines (data augmentation using manual perturbation or GANs) can bring gains in most cases. But on DeepShip, these methods only provide limited and sometimes negative effects (see the underlined part in Table~\ref{tab:main}). This is due to complex factors that affect ship-radiated signals in ocean environments and channels. In such complicated scenarios, manual data augmentation or GANs may not consistently lead to improvements due to the inherent deviations between simulations and real environments. Since our proposed smoothness-inducing regularization prohibits noisy samples from participating in the loss calculation with the ground truth, it could reduce the risk of aggravating misjudgment. By combining data pruning and smoothness-inducing regularization, we achieve a recognition accuracy of up to 88.79\% on Shipsear, which significantly exceeds the original SOTA (85.48\%). On Deepship, we also achieve 80.19\% recognition accuracy, which is a satisfactory result compared to the benchmark.

\begin{figure}[t]
    \centering
    \includegraphics[width=\linewidth]{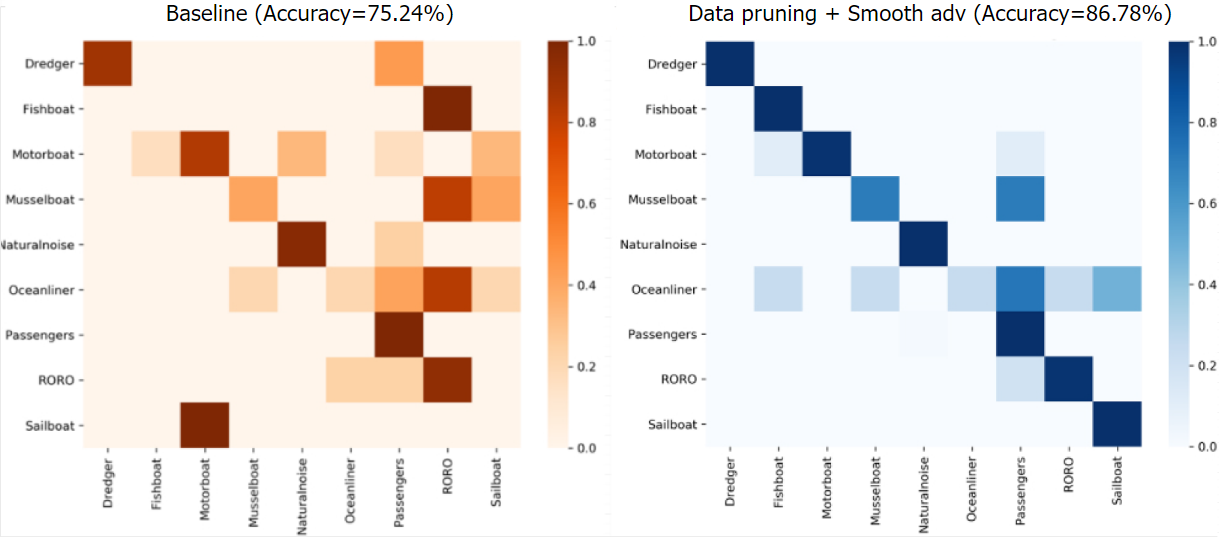}
    \caption{Confusion matrix heat maps on Shipsear. The color scale used in the heatmap ranges from light (low values) to dark (high values).}
    \label{fig:confuse}
\vspace{-2mm}
\end{figure}

Additionally, we make an in-depth analysis of the model's predictions for each category by the confusion matrix heat map. As can be seen from Fig.\ref{fig:confuse}, the baseline model struggles to recognize fish boats and sailboats and often misclassifies other targets as motorboats, mussel boats, or oceanliners. It indicates poor generalization ability and a serious local bias. By applying our techniques, the recognition accuracy for each class is significantly improved, and the model can now recognize fish boats and sailboats that were previously unrecognized. Furthermore, we observe that data pruning greatly reduces the bias of the model towards motorboats and mussel boats, further confirming the effectiveness of data pruning.

\vspace{-3mm}
\subsection{Regularization coefficient $\alpha$}



\begin{figure}[t]
    \centering
    \includegraphics[width=\linewidth]{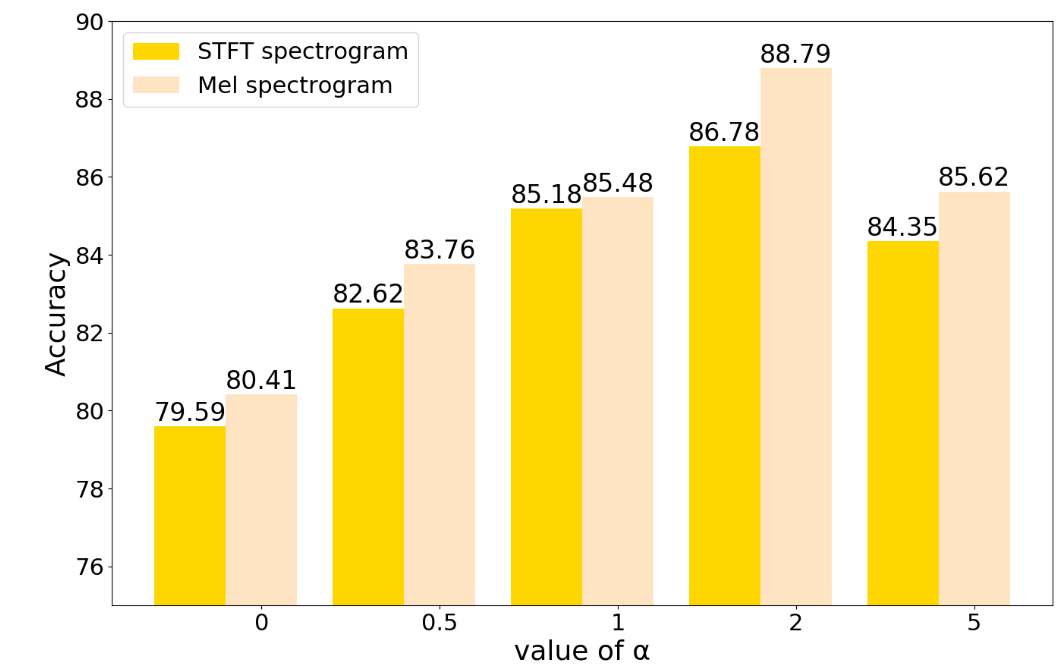}
    \caption{Experiments on Shipsear with different coefficient $\alpha$.}
    \label{fig:bar}
    \vspace{-2mm}
\end{figure}

We explore the influence of the regularization coefficient $\alpha$ on model performance. We could control the weight of the regularization term in the optimization objective by changing $\alpha$. In Fig.\ref{fig:bar}, we observe that  $\alpha$=2 is a good choice. When $\alpha$ is small, the model may still suffer from overfitting. When $\alpha$ is large, the regularization may be too aggressive, thus affecting the learning of the model.

\vspace{-3mm}
\subsection{Efficiency of Data Pruning}

In this subsection, we investigate the optimal pruning thresholds and the corresponding training time reductions for various methods when applying data pruning. Table~\ref{tab:pruning} indicates that data pruning greatly improves recognition performance while reducing training time costs.

Moreover, our findings reveal that different regularization methods require varying optimal pruning thresholds. A high threshold represents an aggressive pruning strategy, which is suitable for dealing with situations where the double-descent phenomenon is serious. Conversely, models that are less prone to local bias necessitate a smaller pruning threshold to prevent the loss of valuable information.

\begin{table}[t]
  \centering
  \caption{Experiments of data pruning on Shipsear. We apply Mel spectrogram as the default feature and conduct experiments under several pruning thresholds (1e-4,1e-5,1e-6). We also report the best results and corresponding thresholds.}
  \resizebox{\linewidth}{!}{
  \begin{threeparttable}
    \begin{tabular}{lcccccc}
    \toprule
    & \multicolumn{3}{c}{Accuracy} & \multicolumn{3}{c}{Training time cost (minutes)}\\
    \cmidrule(lr){2-4}\cmidrule(lr){5-7}
    Methods & prune($\times$)& prune($\checkmark$)  & Best threshold &prune($\times$)& prune($\checkmark$)&Reduction\\
    \midrule
     -  & 77.14 & 80.41 &1e-4& 27.6&12.0 &-56.5\% \\
    GANs  & 77.68  & 82.11 &1e-5 & 27.8 & 18.2 &-34.5\% \\ 
    Manual data aug & 77.72 & 82.40 &1e-4& 48.0& 16.8 &-65.0\%\\
    Smooth-reg & \textbf{82.76}  & \textbf{88.79} &1e-5& 47.6& 42.2 &-11.3\%\\

    \bottomrule
    \end{tabular}
    \end{threeparttable}}
  \label{tab:pruning}%
  \vspace{-3mm}
\end{table}%

\vspace{-2mm}
\section{Conclusions}

In this paper, we point out that the periodic repeatability of ship-radiated signals could lead to the double-descent phenomenon and performance degradation. Our work proposes an adaptive data pruning strategy that dynamically prunes training data to address this issue. Moreover, we propose smoothness-inducing regularization to mitigate overfitting caused by the reduction of training data. Our experiments demonstrate that data pruning is effective and efficient, and our smoothness-inducing regularization outperforms other regularization strategies such as data augmentation. It is noteworthy that our approach can outperform SOTA using simple acoustic features (CQT, Mel spectrograms) and a simple recognition network (ResNet18 with multi-head attention), which leaves ample room for further improvement. In future work, we are interested in exploiting the potentialities of two strategies on more advanced features and recognition networks. We also plan to investigate more fancy approaches to tackle the periodic repeatability problem of ship-radiated signals.

\newpage

\bibliographystyle{IEEEbib}
\bibliography{refs}

\begin{thebibliography}{10}

\bibitem{robertson1995artificial}
James~A Robertson, John~C Mossing, and Bruce~A Weber,
\newblock ``Artificial neural networks for acoustic target recognition,''
\newblock in {\em Applications and Science of Artificial Neural Networks}.
  SPIE, 1995, vol. 2492, pp. 939--950.

\bibitem{heupel2006automated}
Michelle~R Heupel, Jayson~M Semmens, and Alistair~J Hobday,
\newblock ``Automated acoustic tracking of aquatic animals: scales, design and
  deployment of listening station arrays,''
\newblock {\em Marine and Freshwater Research}, vol. 57, no. 1, pp. 1--13,
  2006.

\bibitem{lecun2015deep}
Yann LeCun, Yoshua Bengio, and Geoffrey Hinton,
\newblock ``Deep learning,''
\newblock {\em nature}, vol. 521, no. 7553, pp. 436--444, 2015.

\bibitem{gombos2019acoustic}
Torstein Anton~Berle Gombos,
\newblock ``Acoustic recognition with deep learning; experimenting with data
  augmentation and neural networks,''
\newblock M.S. thesis, 2019.

\bibitem{xie2022underwater}
Yuan Xie, Jiawei Ren, and Ji~Xu,
\newblock ``Underwater-art: Expanding information perspectives with text
  templates for underwater acoustic target recognition,''
\newblock {\em The Journal of the Acoustical Society of America}, vol. 152, no.
  5, pp. 2641--2651, 2022.

\bibitem{perrone1975analysis}
Anthony~J Perrone and Louis~A King,
\newblock ``Analysis technique for classifying wind-and ship-generated noise
  characteristics,''
\newblock {\em The Journal of the Acoustical Society of America}, vol. 58, no.
  6, pp. 1186--1189, 1975.

\bibitem{hernandez2022scaling}
Danny Hernandez, Tom Brown, Tom Conerly, Nova DasSarma, Dawn Drain, Sheer
  El-Showk, Nelson Elhage, Zac Hatfield-Dodds, Tom Henighan, Tristan Hume,
  et~al.,
\newblock ``Scaling laws and interpretability of learning from repeated data,''
\newblock {\em arXiv preprint arXiv:2205.10487}, 2022.

\bibitem{stowell2015acoustic}
Dan Stowell and David Clayton,
\newblock ``Acoustic event detection for multiple overlapping similar
  sources,''
\newblock in {\em 2015 IEEE Workshop on Applications of Signal Processing to
  Audio and Acoustics (WASPAA)}. IEEE, 2015, pp. 1--5.

\bibitem{chen2019integrating}
Hangting Chen, Zuozhen Liu, Zongming Liu, Pengyuan Zhang, and Yonghong Yan,
\newblock ``Integrating the data augmentation scheme with various classifiers
  for acoustic scene modeling,''
\newblock {\em arXiv preprint arXiv:1907.06639}, 2019.

\bibitem{cui2015data}
Xiaodong Cui, Vaibhava Goel, and Brian Kingsbury,
\newblock ``Data augmentation for deep neural network acoustic modeling,''
\newblock {\em IEEE/ACM Transactions on Audio, Speech, and Language
  Processing}, vol. 23, no. 9, pp. 1469--1477, 2015.

\bibitem{gao2020recognition}
Yingjie Gao, Yuechao Chen, Fangyong Wang, and Yalong He,
\newblock ``Recognition method for underwater acoustic target based on dcgan
  and densenet,''
\newblock in {\em 2020 IEEE 5th International Conference on Image, Vision and
  Computing (ICIVC)}. IEEE, 2020, pp. 215--221.

\bibitem{yang2020gan}
Hongbin Yang, Han Gu, Jinyong Yin, and Jian Yang,
\newblock ``Gan-based sample expansion for underwater acoustic signal,''
\newblock in {\em Journal of physics: conference series}. IOP Publishing, 2020,
  vol. 1544, p. 012104.

\bibitem{xie2022adaptive}
Yuan Xie, Jiawei Ren, and Ji~Xu,
\newblock ``Adaptive ship-radiated noise recognition with learnable
  fine-grained wavelet transform,''
\newblock {\em Ocean Engineering}, vol. 265, pp. 112626, 2022.

\bibitem{ren2022ualf}
Jiawei Ren, Yuan Xie, Xiaowei Zhang, and Ji~Xu,
\newblock ``Ualf: A learnable front-end for intelligent underwater acoustic
  classification system,''
\newblock {\em Ocean Engineering}, vol. 264, pp. 112394, 2022.

\bibitem{gong2021eliminate}
Yunpeng Gong, Liqing Huang, and Lifei Chen,
\newblock ``Eliminate deviation with deviation for data augmentation and a
  general multi-modal data learning method,''
\newblock {\em arXiv preprint arXiv:2101.08533}, 2021.

\bibitem{huber2011robust}
Peter~J Huber,
\newblock ``Robust statistics,''
\newblock in {\em International encyclopedia of statistical science}, pp.
  1248--1251. Springer, 2011.

\bibitem{jiang2019smart}
Haoming Jiang, Pengcheng He, Weizhu Chen, Xiaodong Liu, Jianfeng Gao, and Tuo
  Zhao,
\newblock ``Smart: Robust and efficient fine-tuning for pre-trained natural
  language models through principled regularized optimization,''
\newblock {\em arXiv preprint arXiv:1911.03437}, 2019.

\bibitem{santos2016shipsear}
David Santos-Dom{\'\i}nguez, Soledad Torres-Guijarro, Antonio
  Cardenal-L{\'o}pez, and Antonio Pena-Gimenez,
\newblock ``Shipsear: An underwater vessel noise database,''
\newblock {\em Applied Acoustics}, vol. 113, pp. 64--69, 2016.

\bibitem{irfan2021deepship}
Muhammad Irfan, Zheng Jiangbin, Shahid Ali, Muhammad Iqbal, Zafar Masood, and
  Umar Hamid,
\newblock ``Deepship: An underwater acoustic benchmark dataset and a separable
  convolution based autoencoder for classification,''
\newblock {\em Expert Systems with Applications}, vol. 183, pp. 115270, 2021.

\bibitem{kingma2014adam}
Diederik~P Kingma and Jimmy Ba,
\newblock ``Adam: A method for stochastic optimization,''
\newblock {\em arXiv preprint arXiv:1412.6980}, 2014.

\bibitem{he2016deep}
Kaiming He, Xiangyu Zhang, Shaoqing Ren, and Jian Sun,
\newblock ``Deep residual learning for image recognition,''
\newblock in {\em Proceedings of the IEEE conference on computer vision and
  pattern recognition}, 2016, pp. 770--778.

\bibitem{Wang2018NonlocalNN}
X.~Wang, Ross~B. Girshick, Abhinav~Kumar Gupta, and Kaiming He,
\newblock ``Non-local neural networks,''
\newblock in {\em CVPR}, 2018.

\end{thebibliography}

\end{document}